\documentclass[12pt]{iopart}
\usepackage{graphicx}
\usepackage[font=small,labelfont=bf,
   justification=justified,
   format=plain]{caption}
\usepackage{subcaption}
\usepackage{tikz}
\usetikzlibrary{calc}
\usepackage{hyperref}
\usepackage{wrapfig}
\usepackage{svg}
\usepackage{hhline}
\usepackage{diagbox}
\usepackage{adjustbox}
\usepackage{algorithm}
\usepackage[ruled,vlined,algo2e]{algorithm2e}

\usepackage[symbol]{footmisc}

\usepackage{soul}

\soulregister\cite7
\soulregister\eqref7

\usepackage{lineno}
\begin{document}


\title{An Efficient Convex Hull-Based Vehicle Pose Estimation Method for 3D LiDAR}

\author{Ningning Ding$^{1,*}$, Ruihao Ming$^1$, and Bo Wang$^1$.}


\address{Jiangsu Jinling Intelligent Manufacturing Research Institute Co., Ltd., Nanjing, China.}

\ead{dingnn@mail.ustc.edu.cn}
\vspace{0.1 in}

\begin{abstract}

Vehicle pose estimation with LiDAR is essential in the perception technology of autonomous driving. However, due to incomplete observation measurements and sparsity of the LiDAR point cloud, it is challenging to achieve satisfactory pose extraction based on 3D LiDAR with the existing pose estimation methods. In addition, the demand for real-time performance further increases the difficulty of the pose estimation task. In this paper, we propose a novel vehicle pose estimation method based on the convex hull. The extracted 3D cluster is reduced to the convex hull, reducing the subsequent computation burden while preserving essential contour information. Subsequently, a novel criterion based on the minimum occlusion area is developed for the search-based algorithm, enabling accurate pose estimation. Additionally, this criterion renders the proposed algorithm particularly well-suited for obstacle avoidance. The proposed algorithm is validated on the KITTI dataset and a manually labeled dataset acquired at an industrial park. The results demonstrate that our proposed method can achieve better accuracy than the classical pose estimation method while maintaining real-time speed.

\end{abstract}

\vspace{2pc}

\submitto{TRR}

%
%
\newpage
\section{Introduction}

Autonomous driving has drastically developed in the last decades. The safety of autonomous driving requires a reliable perception system to perceive surroundings for mobile platforms. In environmental perception technology, 3D object detection is one of its primary research directions, and researchers have applied various sensors to accomplish this task \cite{li2022bevdepth,wang2021plumenet,ma20223d,zamanakos2021comprehensive}. In particular, the 3D Light Detection and Ranging (LiDAR) sensor has gained widespread deployment due to its numerous advantages, including all-weather operation, centimeter-level accuracy, and the ability to measure great distances compared with stereo cameras \cite{behley2019semantickitti}. This article involves LiDAR-based 3D object detection, which can be categorized into two main types: traditional methods and deep learning methods. In recent years, various learning-based 3D object detection methods have been proposed, including VoxelNet \cite{zhou2018voxelnet}, PointRCNN \cite{shi2019pointrcnn}, PointPillars \cite{lang2019pointpillars}, CenterPoint \cite{yin2021center}, among others. Most of these approaches offer an end-to-end solution for 3D object detection using a 3D convolutional neural network, showcasing exceptional performance. Nevertheless, there exist two inherent potential limitations. Firstly, learning-based methods necessitate human labeling, a laborious and time-consuming task. Secondly, the performance of learning-based 3D object detection methods may degrade when applied to scenes that differ from the training dataset or have distinct sensor configurations. In contrast, traditional methods demonstrate superior adaptability. In the conventional 3D LiDAR-based object detection pipeline, the typical procedure involves initial ground segmentation of the original 3D point cloud, followed by point cloud clustering of the non-ground points. The final step entails estimating the object's bounding box based on the clustering results, thereby accomplishing the 3D object detection task. The main advantage of the traditional methods is that they do not need prior information about the environment for training.

The work in this paper focuses on estimating the vehicle's pose with the clustering point cloud in a LiDAR perception pipeline described above. The main contributions of this paper are listed as follows:

\begin{itemize}
    \item We developed a convex hull-based vehicle pose estimation method for 3D LiDAR. A novel criterion based on the minimum occlusion area is employed, which can achieve better performance while maintaining real-time speed. Moreover, our proposed method is particularly suitable for path planning due to the minimum occlusion area criterion.
    
    \item Our proposed algorithm was evaluated with the KITTI dataset \cite{geiger2013vision} and our own dataset. Experimental evidence corroborates the proposed algorithm exhibits promising performance compared with state-of-the-art methods.
\end{itemize}

\section{Related work} 
The existing vehicle pose estimation algorithms for 3D LiDAR based on traditional methods can be categorized into two types: feature-based methods and global pose estimation algorithms. The feature-based methods rely on the edge features to infer the pose of the objects \cite{xu2018real,lin2021adaptive,zhao2021shape}. \hl{The distribution shape of the vehicular point cloud is generally classified into L-shape, I-shape, and C-shape.\cite{wittmann2014improving}} Considering the L-shape contour frequently appears in the distribution of 3D LiDAR point clouds \cite{zhao2021shape}, many methods have used this feature for object pose estimation. Zhang \textit{et al} \cite{zhang2017efficient} proposed a search-based method for pose estimation of the vehicle. It iterates all possible directions of the rectangle, and multiple criteria are applied to evaluate the optimal fitting directions. However, this method is computationally expensive and faces challenges in achieving real-time speed when employed to fit numerous objects simultaneously. Qu \textit{et al} \cite{qu2018efficient} innovatively decomposes the L-shape fitting problems into two distinct steps: L-shape vertex searching and L-shape corner point locating using 2D laser data. This decomposition effectively reduces the computational complexity. \hl{Yang \cite{liu2020estimation} proposed an efficient bounding box estimation method for 2D bird’s-eye view LiDAR points. This method evaluates each candidate angle formed by non-adjacent convex hull points and selects the orientation that can optimize a given criterion. Although the algorithm efficiency is greatly improved compared to the original L-shape fitting method, the accuracy is not ideal due to the criterion.} Kim \textit{et al} \cite{kim2017shape}] introduced an iterative end-point fitting method for extracting L-shape features from the 3D point cloud. This method begins by designating the minimum and maximum clustering angle points as the baseline's endpoints. The break-point is determined as the point that exhibits the maximum distance from the baseline to each cluster point. The final pose estimation bounding box is established by identifying the farthest point from the baseline as the L-shaped corner point. However, this method assumes the presence of a complete L-shape in all point clouds. To overcome this limitation, Zhao \textit{et al} \cite{zhao2021shape} proposed a new corner edge-based L-shape fitting method. This approach involved computing an initial approximation of the corner point and subsequently distinguishing whether it represented the corner point of the L-shape or the side point of the vehicle. The RANSAC algorithm was then employed to fit the L-shape features and get the pose estimation result. \hl{It should be noted that this method may degenerate when the vehicle's point cloud is not L-shaped or I-shaped.}


In comparison to the feature-based methods, global-based pose estimation algorithms rely less on the specific distribution of the point cloud and instead emphasize the global applicability of the algorithm. In reference \cite{ naujoks2018orientation}, the authors proposed an orientation-corrected bounding box fitting method based on the convex hull and a line creation heuristic. This method successfully achieves correct pose estimation, even in scenarios involving sparse point clouds, by capitalizing on the efficiency of convex hull extraction. Liu \textit{et al} \cite{liu2019fast} introduced a pose estimation method that relies on the convex-hull model and position inference. The bounding box’s orientation is determined by multiple evaluation criteria. Yang \textit{et al} \cite{yang2019vehicle} introduced a novel vehicle pose estimation method based on edge distance. The introduced edge distance effectively captures the point cloud distribution. Subsequently, a bounding rectangle is derived from the edge distance to estimate the vehicle's pose. Additionally, if the vehicle exhibits only one visible side or the point clouds are sparse, a correction mechanism is employed to refine the bounding rectangle and obtain a more accurate vehicle pose estimation. An \textit{et al} \cite{an2020novel} proposed a vehicle pose estimation algorithm utilizing a low-end 3D LiDAR. The algorithm initially constructed four vehicle models with distinct observation angles, subsequently representing the vehicle's measured size as a uniformly distributed sample. Finally, the algorithm employed template matching principles to estimate the vehicle's pose. He \textit{et al} \cite{he2022pose} proposed a pose estimation method for moving vehicles based on the heuristic rules. This method utilizes heuristic rules that leverage edge length and edge visibility to filter candidate orientations generated by the convex hull sampling method. Subsequently, tracking heuristics are applied to enhance the smoothness of the orientation estimation results. \hl{However, the performance of this method is limited to the non-ideal criterion.} Sun \textit{et al} \cite{sun2023dynamic} presents a dynamic vehicle pose estimation method utilizing heuristic L-shape fitting and a grid-based particle filter. A geometric shape classifier module categorizes clusters into symmetrical and asymmetrical ones. For asymmetrical clusters, a contour-based heuristic L-shape fitting module is introduced, while the pose estimation of symmetrical clusters is performed using a structure-aware grid-based particle filter. \hl{Jin \textit{et al} \cite{jin2023robust} adopts different fitting strategies for vehicle contour points in various states to improve the algorithm's adaptability. Moreover, the principal component analysis method is introduced into the line-fitting process to enhance the robustness of the bounding box fitting results. In reference \cite{xu2023dynamic}, the heading normalization vehicle model and a robust vehicle size estimation method are introduced to estimate the vehicle pose with 2D-matched filtering.}

\hl{Most feature-based methods specialize in targets with specific point cloud distribution, such as the L-shape and I-shape. However, their performances may degenerate sharply when dealing with other types of point cloud distribution which is often the case for real-world traffic scenes. Although the global-based methods have better
adaptability than feature-based methods in complex scenes, they are often computationally expensive, especially for large datasets, and may require necessitating parameter tuning to achieve optimal results. To overcome these constraints, there is an urgent demand for a robust and efficient vehicle pose estimation method capable of simultaneously handling targets with diverse point cloud distribution while minimizing computational overhead.}

\begin{figure*}
\centering
\subfloat[]{\label{flow_chart-a} \includegraphics[scale=0.15]{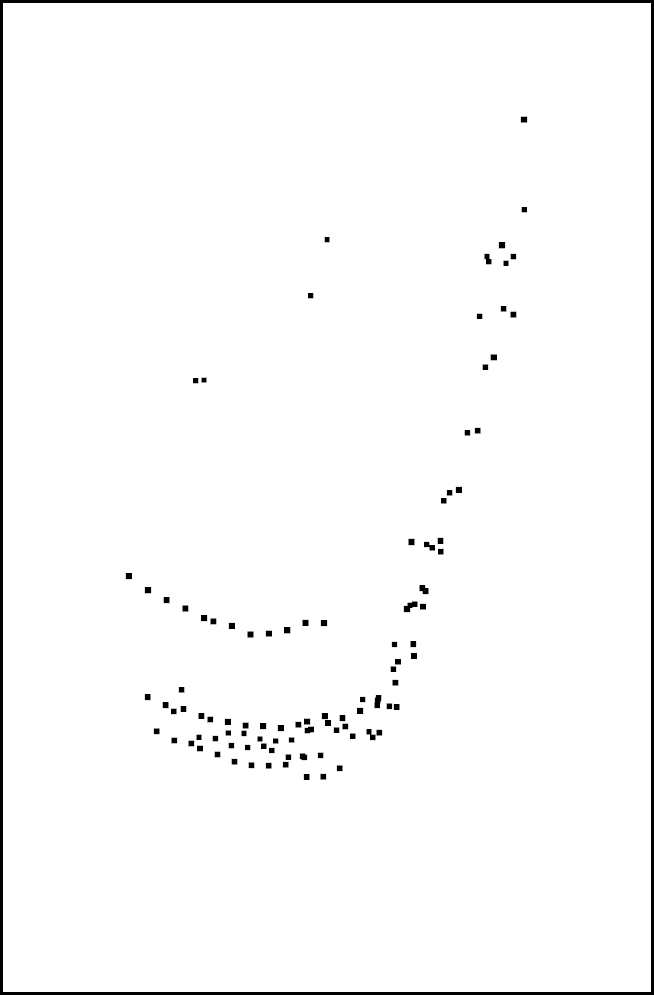}}
\hfill
\subfloat[]{\label{flow_chart-b} \includegraphics[scale=0.15]{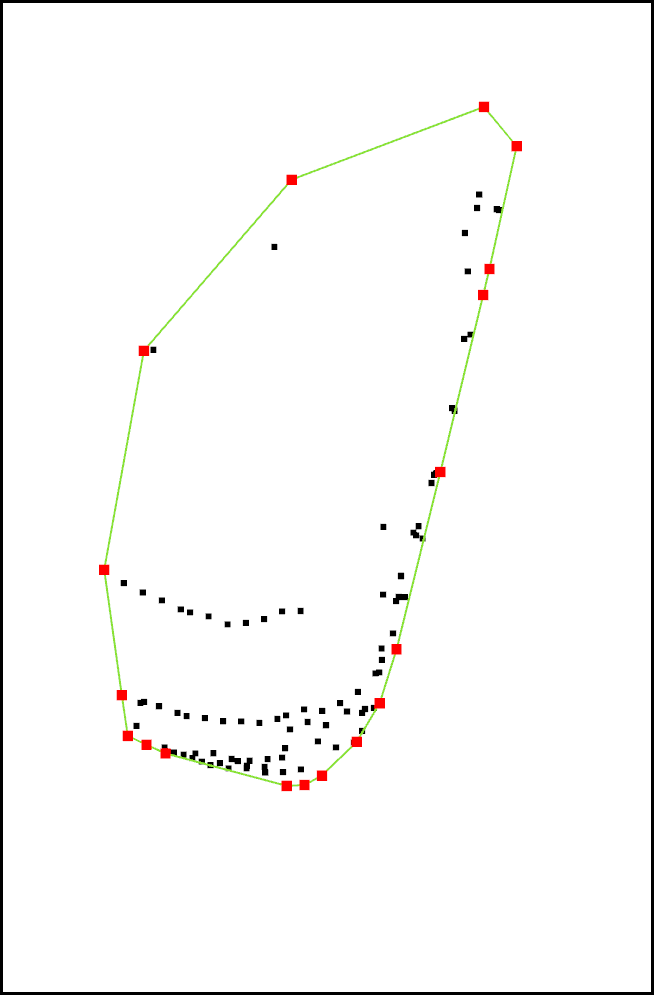}}%
\hfill
\subfloat[]{\label{flow_chart-c} \includegraphics[scale=0.15]{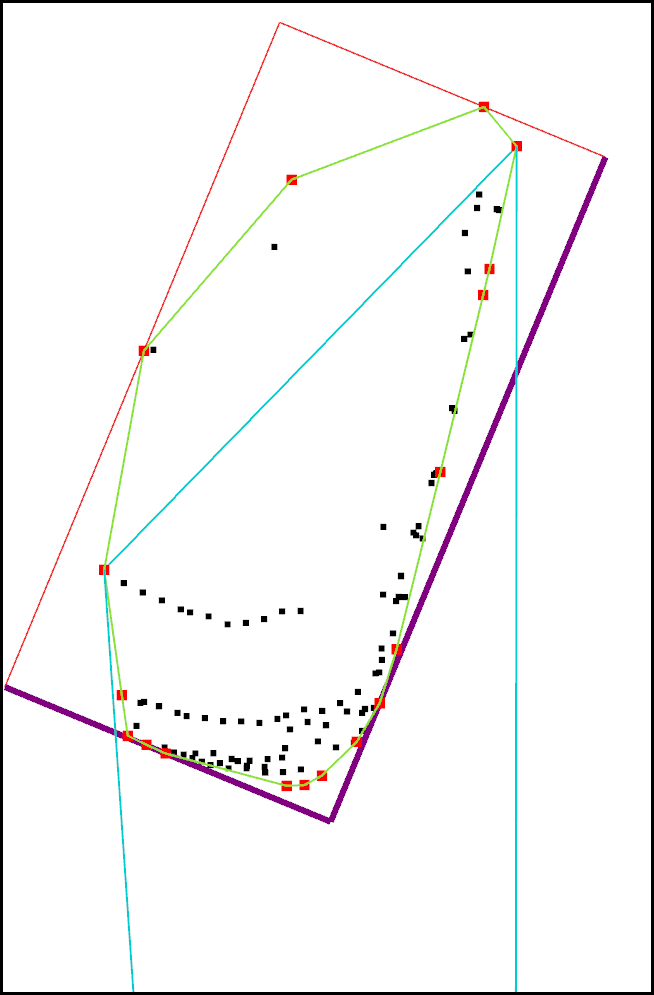}}%
\hfill
\subfloat[]{\label{flow_chart-d} \includegraphics[scale=0.15]{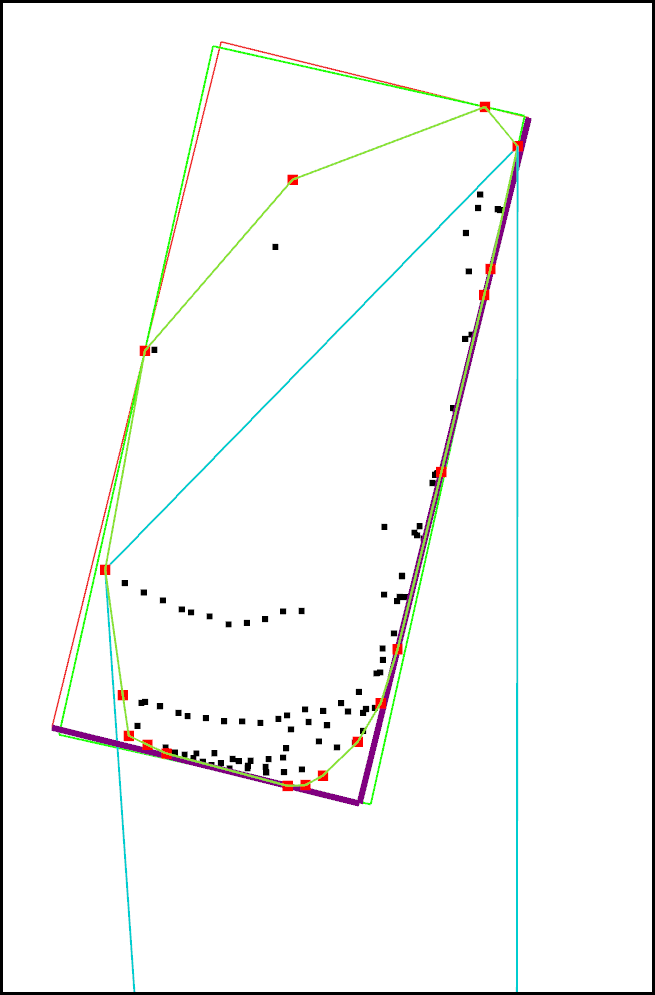}}%
\caption{\hl{Illustration of the proposed method. (a) A raw 3D point cloud cluster. (b) The extracted convex hull points. (c) Visible edge for a certain pose (The purple lines), the occlusion region's area will be calculated. (d) The final fitting result after optimization. (The red bounding box is obtained by our method. The green bounding box is the ground truth result.)}}
\label{fig:flow_chart}
\end{figure*}

\section{Methodology}
The proposed method consists of several key steps to evaluate a candidate orientation for the vehicle, including convex hull extraction, visible edge selection, and occlusion area calculation. In this section, we will discuss the detailed implementation of the proposed algorithm, as well as its fundamentals. The overall illustration of the proposed pose estimation algorithm is shown in Figure \ref{fig:flow_chart}.

Since we only care about the yaw angle of objects in the autonomous \hl{drivering} application, we can project the 3D point cloud data into the x-y plane. This reduces the original clusters from 3D space to 2D space and reduces the computational burden for the following steps. However, the projected 2D cluster lost the information along the z-axis. To get the height and position of the bounding box along the z-axis, we should calculate the minimum and maximum z values of the 3D cluster, namely $z_{min}$ and $z_{max}$, before projection. The height of the bounding box equals $z_{max}-z_{min}$, and the position of the bounding box along the z-axis is $(z_{max}-z_{min})/2$. The aim of this work is to improve speed while ensuring pose estimation accuracy. So we choose to exact the contour from the cluster and avoid dealing with point clouds directly. To this end, we can extract the convex-hull points from the projected 2D cluster. For consideration of both simplicity and efficiency, we choose the Graham scan method \cite{graham1972efficient} to generate a set of convex-hull points from the projected 2D point cloud.

Then, we use a search-based algorithm to find the optimal vehicle pose. The key issue of this algorithm is to choose the proper criterion to decide which orientation is the best option. There are already many criteria proposed in the literature, like area minimization, closeness maximization, and variance minimization \cite{zhang2017efficient}. However, these criteria are only appropriate for the 2D cluster, and its performance will degrade when using a convex hull. This is due to some information being lost when reducing the 2D cluster into a convex hull. Besides, these criteria are mainly suitable when the cluster is an L-shape. However, due to occlusion and different viewing angles, the L-shape cannot be observed in some cases.

\begin{algorithm}[H]\footnotesize
    \caption{Pipeline of the Proposed Algorithm}\label{alg:pipeline_proposed}
    \SetAlgoLined
    \SetKwInOut{Input}{Input}\SetKwInOut{Output}{Output}
    \Input{\hl{cluster}'s point cloud $cluster\_pc$}
    \Output{orientation estimation value ${\theta }^{*}$}
    \BlankLine
    \hspace{0.01cm}$occluAreaArray \leftarrow$ an empty vector;\\
    $cluster\_pc\_2d \leftarrow projection2D(cluster\_pc)$\\
    $ch\_pts \leftarrow convexHull(cluster\_pc\_2d)$\\
    \For{$i$ \textbf{in} $[0,ch\_pts.size())$}{
        $pt\_azimuth.push\_back( atan2(ch\_pts[i].y, ch\_pts[i].x))$;
    }
    \hspace{0.01cm}$vl\_idx\_l=argmin_i(pt\_azimuth[i])$;\\
    $vl\_idx\_r=argmax_i(pt\_azimuth[i])$;\\
    \For{$\theta$ \textbf{in} $[0,\pi/2]$ \textbf{step} $\delta$}{
        \For{$i$ \textbf{in} $[0,4)$ }{
            \hspace{0.01cm}${\vec{e}}_{1}\leftarrow [cos \theta ,sin \theta ]$, ${\vec{e}}_{2}\leftarrow [-sin \theta ,cos \theta ]$;\\
            ${{C}_{1}}\leftarrow Pt\cdot {{\vec{e}}_{1}}$, ${{C}_{2}}\leftarrow Pt\cdot {{\vec{e}}_{2}}$;\\
            $line[0].a \leftarrow cos \theta,\ line[0].b \leftarrow sin \theta,\ line[0].c \leftarrow \min \{{{C}_{1}}\}$;\\
            $line[1].a \leftarrow -sin \theta,\ line[1].b \leftarrow cos \theta,\ line[1].c \leftarrow \min \{{{C}_{2}}\}$;\\       
            $line[2].a \leftarrow cos \theta,\ line[2].b \leftarrow sin \theta,\ line[2].c \leftarrow \max \{{{C}_{1}}\}$;\\   
            $line[3].a \leftarrow -sin \theta,\ line[3].b \leftarrow cos \theta,\ line[3].c \leftarrow \max \{{{C}_{2}}\}$; 
        }

        \hspace{0.01cm}$[projL\_idx\_l, projL\_idx\_r] = Detect Visible Edge()$;\\
        $ [\theta,occluArea]=Calculate Occlusion Area()$;\\
        $occluAreaArray.push\_back([\theta,occluArea])$
    }
   ${\theta }^{*} = occluAreaArray[argmin_i(occluAreaArray[i].occluArea)].\theta$;
\end{algorithm}

In this paper, we proposed a novel criterion based on the minimum occlusion area or maximum driveable area. This is a straightforward idea since the wrong estimation of vehicle orientation will result in the driveable region being occupied by mistake. A schematic illustration of this phenomenon is shown in Figure \ref{fig:schematic_diagram}. \hl{The rectangle is divided into three parts: the convex hull region, occlusion region (OR), and invalid region (IR)}. The goal of our criterion is to minimize this wrong occlusion region. It should be noted that the area of OR is hard to calculate if using the original 2D cluster. Fortunately, this becomes easy after converting the 2D cluster to a convex hull. The pipeline of the proposed algorithm is presented in Alg.\ref{alg:pipeline_proposed}.

\subsection{Visible edge selection}
\begin{figure}
\centerline{\includegraphics[scale=0.20]{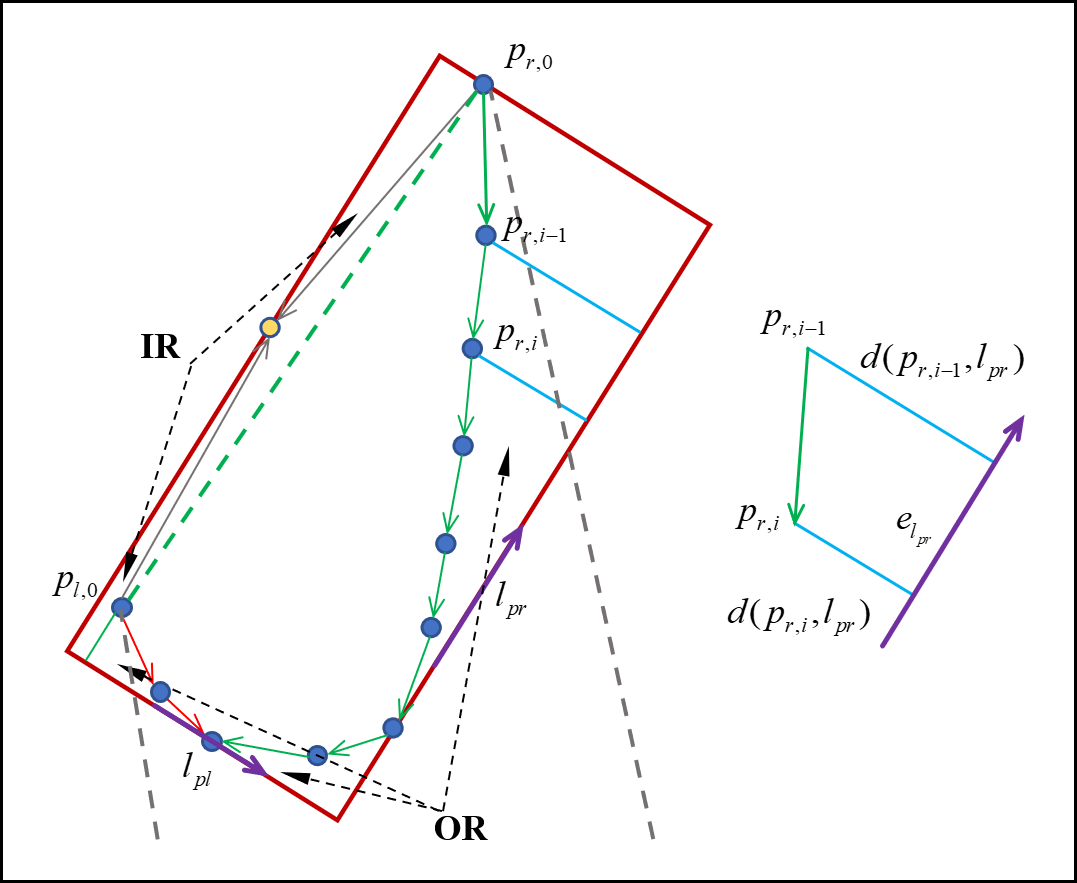}}
\caption{Schematic illustration of the visible edge selection and occlusion area calculation.
\label{fig:schematic_diagram}}
\end{figure}

Inspired by the concept of calculus, the area of OR can be divided into many trapezoids, as shown in Figure \ref{fig:schematic_diagram}. To calculate its area, the first thing is to choose the appropriate projection edge from the rectangle's four edges so that we can calculate the trapezoid's height.

\begin{figure}[h]
   	\centering
    \subfloat[\label{0a}]{
        \includegraphics[scale=0.21]{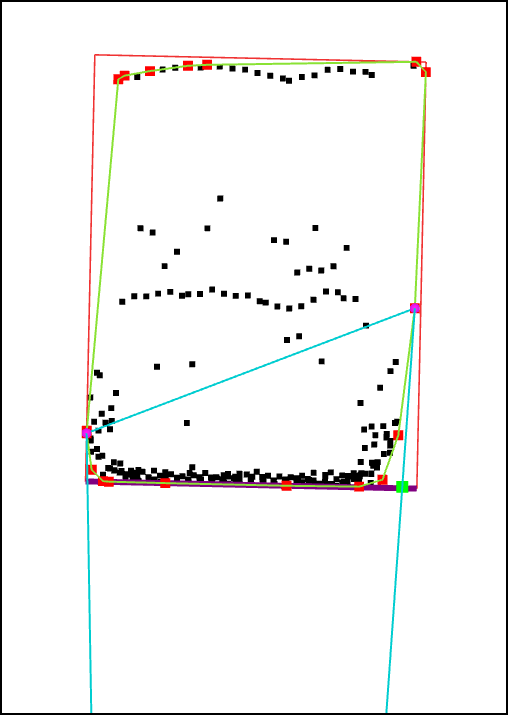}}
    \subfloat[\label{0b}]{    
        \includegraphics[scale=0.2105]{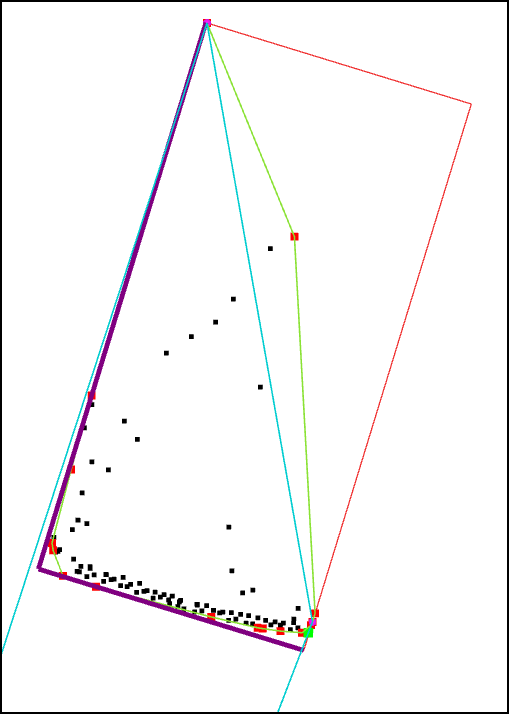}}
	\caption{Sample results of the projection edge selection (The purple line denotes the projection edge). (a) Only one projection edge. (b) Two projection edges.}
	\label{fig:demo_proj_line}
\end{figure}

Considering the vehicle body facing away from the LiDAR can't be detected, the rectangle's edges here have no valid information to decide the vehicle's pose. Therefore, we choose the visible edges as the projection line. Firstly, we calculate the two boundary points of convex hull points according to the azimuth angle. Then we connect the two border points with the LiDAR origin to form a valid region as shown in the green triangle in Figure \ref{fig:schematic_diagram}. Only points and edges in this region will be considered for vehicle pose estimation. The edge contacts with border lines first will be selected as the projection edge. We first calculate the intersection of border lines and rectangle edges. If the intersection is on the rectangular edge, this edge is chosen as the candidate projection edge. If the intersection coincides with a rectangular vertice, two adjacent rectangular edges will be chosen as the candidate projection edge. Then in all the candidate edges, the one whose intersection is closest to the LiDAR is the projection edge. If the closest intersection happens to be the rectangular vertice, the edge located in the valid region will be the projection edge. The detail of this process is presented in Alg. \ref{alg:visible_edge_det}. \hl{Variable $is\_recv\_$ denotes whether the boundary point is also the rectangular vertice. Whether the boundary line intersects with the rectangular edge is by judging whether the maximum distance between the intersection points and the adjacent rectangular vertices is smaller than the length of the edge.} Figure \ref{fig:demo_proj_line} shows that our method can detect projection edges accurately in different cases.

\subsection{Occlusion area calculation}

\begin{algorithm}[t]\footnotesize
    \caption{Visible Edge Selection}\label{alg:visible_edge_det}
    \SetAlgoLined
    \SetKwInOut{Input}{Input}\SetKwInOut{Output}{Output}
    \Input{Rectangle edges, vertexes, and boundary lines}
    \Output{Projection edges index}
    \BlankLine
    \hspace{0.02cm}$projL\_idex\_l=\textbf{SelectProjLine}(vl\_idx\_l,line\_vl\_l,is\_recv\_l,recv\_idx\_l)$;\\
    $projL\_idex\_r=\textbf{SelectProjLine}(vl\_idx\_r,line\_vl\_r,is\_recv\_r,recv\_idx\_r)$;
    
    \setcounter{AlgoLine}{0}
    \SetKwProg{myproc}{Procedure}{\hspace{0.1cm}$\textbf{SelectProjLine}(vl\_idx\_,line\_vl\_)$}{end}
    \myproc{}{
    \hspace{0.02cm}$candi\_pL\_idx \leftarrow$ an empty vector;\\
    $candiIdx\_array \leftarrow$ an empty vector;\\
    \For{$idx\_$ \textbf{in} $[0,4)$ }{
        \If{$(is\_vertex\_ \ \textbf{and}\  idx\_ == \{recv\_idx\_ \ \textbf{or}\ recv\_idx\_ + 1\})$}{       
            $continue$;
        }
        \hspace{0.01cm}$vl\_poi=calcuIntersection (line[i],line\_vl\_)$;\\
        $vl2rec\_di\_l=dst(vl\_poi,rcr\_pts[idx\_])$;\\    $vl2rec\_di\_r=dst(vl\_poi,rcr\_pts[idx\_ - 1])$;\\
        $edge\_len = dst(rcr\_pts[idx\_-1],rcr\_pts[idx\_])$;\\
        \If{$(max(vl2rec\_di\_l, vl2rec\_di\_r) < edge\_len)$}{
            $candi\_pL\_idx.push\_back(idx\_)$;
        }

        \If{$(\textbf{not}\  is\_recv\_)$}{                    
            \hspace{0.01cm}$ dst\_i = abs(max(vl2rec\_di\_l, vl2rec\_di\_r) - edge\_len)$;\\
            $candiIdx\_array[idx\_].idx\_ = idx\_$;\\
            $candiIdx\_array[idx\_].dst\_ = dst\_i$;
        }
    }
    \hspace{0.01cm}$sort(candiIdx\_array)$;\\
    \If{$(\textbf{not}\ is\_recv\_ \ \textbf{and}\  candi\_pL\_idx.empty())$}{
        \hspace{0.01cm}$candi\_pL\_idx.push\_back(candiIdx\_array[0].idx\_)$;\\
        $candi\_pL\_idx.push\_back(candiIdx\_array[1].idx_)$;\\
        $is\_recv\_ = true$;\\
        $recv\_idx\_ = min(candiIdx\_array[0].idx\_, candiIdx\_array[1].idx\_)$;
    }

    \For{$i$ \textbf{in} $[0,candi\_pL\_idx.size())$ }{
        \hspace{0.01cm}$can\_vl\_poi=calcuIntersection (line[candi\_pL\_idx[i]], line\_vl\_)$;\\
        $len\_can\_vl\_poi.push\_back(len(can\_vl\_poi))$;
    }

    \eIf{$(is\_recv\_ \ \textbf{and}\  len(rcr\_pts[recv\_idx\_]) < min\{len\_can\_vl\_poi\})$}{
        \eIf{$isVisible(rcr\_pts[recv\_idx\_])$}{
            $projL\_idex\_=recv\_idx\_$;
        }
        {
            $projL\_idex\_=recv\_idx\_+1$;
        }
    }{
        \hspace{0.01cm}$projL\_idex\_=argmax_i(len\_can\_vl\_poi[i])$;
    }

    \Return{$projL\_idex\_$}
    }
\end{algorithm}

After determining the projection edge, we can calculate the trapezoid's height and further get the whole OR's area. We start from the right boundary point. There are two different directions to search for the next point. The adjacent point in the visible region will be selected as the next point. We will continue to search the rest of the points in this direction, as shown in Figure \ref{fig:schematic_diagram}. The two adjacent points along the right projection edge form a trapezoid. For the i-th trapezoid, its area can be calculated as

\begin{equation}
 {{OR}_{r,i}}=[ d({{p}_{r,i}},{{l}_{pr}})+d({{p}_{r,i-1}},{{l}_{pr}})]\cdot(\overrightarrow{{{p}_{r,i}}{{p}_{r,i-1}}} \cdot \overrightarrow{{{e}_{{{l}_{pr}}}}} )/2
\label{eq:box_update}
\end{equation}

\noindent where $d({{p}_{r,i}},{{l}_{pr}})$ denotes the relative distance of ${{p}_{r,i}}$ to the projection line ${l}_{pr}$, $\overrightarrow{{{e}_{{{l}_{pr}}}}}$ is the unit direction vector of projection line ${l}_{pr}$.

To ensure the trapezoid is in the visible region, we should give a stopping criterion to judge whether the calculated trapezoid is valid and stop the next calculation. We first calculate the projection of $\overrightarrow{{{p}_{r,i-1}}{{p}_{r,i}}}$ and $\overrightarrow{{{p}_{r,i-2}}{{p}_{r,i-1}}}$ on projection edge ${l}_{pr}$. If the two projection vectors are in the opposite direction, it means that the current trapezoid is not valid, and the \hl{calculation along this direction} is terminated. This stopping criterion can be simplified by calculating the product of $\overrightarrow{{{p}_{r,i-1}}{{p}_{r,i}}} \overrightarrow{{{e}_{{{l}_{pr}}}}}$ and  $\overrightarrow{{{p}_{r,i-2}}{{p}_{r,i-1}}} \overrightarrow{{{e}_{{{l}_{pr}}}}}$. If the sign of this product is negative, the stopping criterion is activated. If the last valid point happens to be the left boundary point, it means the occlusion area calculation has been completed. If the last valid point is not the left boundary point, we need to search for the valid trapezoid from the left boundary point until it reaches the breakpoint. Finally, we can get the area of the occlusion region when the fitted rectangle is in a certain orientation. The detail of this process is presented in Alg. \ref{alg:occul_area_calcu}. \hl{The variable $iter\_dir\_$ is used to determine the search direction of the initial boundary points. The variable $gap\_num\_$ is used to record the number of points from the break position to another boundary point.} 

It should be noted that the possible orientation of the fitted rectangle $\theta$ ranges from $0^{\circ}$ to $90^{\circ}$, as the two consecutive sides of a rectangle are orthogonal and we only consider the single edge falling within $0^{\circ}$ and $90^{\circ}$. Thus, we can traverse all possible orientations of the rectangle with a fixed resolution. Then we calculate the corresponding occlusion region’s area and choose the orientation with the minimum occlusion area as the estimated orientation of our algorithm. Once we determine the rectangle's orientation, its four edges can be calculated easily, and the detail can be found in reference \cite{zhang2017efficient}. Combining the bounding box's position and dimension along the z-axis, we can finally get the 3D bounding box for an object using our algorithm.

\begin{algorithm}[H]\footnotesize
	\caption{Occlusion Area Calculation}\label{alg:occul_area_calcu}
	\SetAlgoLined
	\SetKwInOut{Input}{Input}\SetKwInOut{Output}{Output}
        \Input{Projection edge index}
	\Output{Area of the occlusion region}
	\BlankLine
	
        \hspace{0.01cm}$iter\_dir\_l \leftarrow 1$; $iter\_dir\_r \leftarrow 1$;\\
        \If{$vl\_idx\_l + 1 == vl\_idx\_r \ \textbf{or} \ isVisible(ch\_pts[vl\_idx\_l-1])$}{$iter\_dir\_l=-1$;}
        \If{$vl\_idx\_r + 1 == vl\_idx\_l \ \textbf{or} \ isVisible(ch\_pts[vl\_idx\_r-1])$}{$iter\_dir\_r=-1$;}
        \hspace{0.01cm}$[occulsion\_area\_r,gap\_num\_r]=\textbf{CalculateOR}(projL\_idx\_r,vl\_idx\_r,\\vl\_idx\_l,iter\_dir\_r,ch\_pts.size)$;\\      
        $[occulsion\_area\_l,gap\_num\_l]=\textbf{CalculateOR}(projL\_idx\_l,vl\_idx\_l,\\vl\_idx\_r, iter\_dir\_l,gap\_num\_r)$;\\
        $occulsion\_area=occulsion\_area\_l+occulsion\_area\_r$;\\  

    \setcounter{AlgoLine}{0}
    \SetKwProg{myproc}{Procedure}{\hspace{0.1cm}$\textbf{CalculateOR}(projL\_idx\_,vl\_idx\_,vl\_idx\_n,iter\_dir\_,iter\_max)$}{end}
    \myproc{}{
        \hspace{0.01cm}$occulsion\_area\_ \leftarrow 0$;\\
        $gap\_num\_ \leftarrow 0$;\\
        $idx \leftarrow vl\_idx\_$;\\
        \hl{$break\_condition \leftarrow 1$;}\\
        \For{$iter\_num$ \textbf{in} $[0,iter\_max)$ }{
            \hspace{0.01cm}$line\_vector = \left[line[projL\_idx\_].b, -line[projL\_idx\_].a\right]$;\\
            $pt\_c = [ch\_pts[idx].x, ch\_pts[idx].y]$;\\
            $pt\_n = [ch\_pts[idx+iter\_dir\_].x, ch\_pts[idx+=iter\_dir\_].y]$;\\
            $ch\_lineseg = pt\_n - pt\_c$;\\
            $trapezoid\_h = ch\_lineseg\cdot line\_vector$;\\
            $idx\_last\_ = idx$;\\
            \If{$idx \ne vl\_idx\_ \ \textbf{and}\  trapezoid\_h \cdot break\_condition < 0$}{$break$;}
            \If{$idx == vl\_idx\_n$}{$break$;}
            \hspace{0.01cm}$trapezoid\_u = pt2linedst(pt\_c, line[projL\_idx\_])$;\\
            $trapezoid\_l = pt2linedst(pt\_n, line[projL\_idx\_])$;\\
            $trapezoid\_area = (trapezoid\_u + trapezoid\_l) \cdot trapezoid\_h / 2$;\\
            $occulsion\_area\_ += trapezoid\_area$;\\
            $break\_condition = trapezoid\_h$;\\
            $idx+=iter\_dir\_$;
        }
        \While{$idx\_last\_ \ne vl\_idx\_n$}
        {
          \hspace{0.01cm}$idx\_last\_ += iter\_dir\_$;\\
          $gap\_num\_++$;
        }
        \Return{$[occulsion\_area\_,gap\_num\_]$}
    }
\end{algorithm}

\section{Experiments}
To evaluate the correctness and efficiency of our proposed algorithm, experiments are carried out on the KITTI dataset and a manually labeled dataset. We compare the proposed algorithm with classical vehicle pose estimation methods like RANSAC L-shape fitting method \cite{zhao2021shape} and convex hull heuristic (CHH) method \cite{he2022pose}. Since our method is not associated with any tracking algorithm, the vanilla CHH without tracking is used here for a fair comparison. These algorithms were all implemented in C++ \hl{framework on the Ubuntu 20.04 system}, and there was no parallelization or GPU acceleration enabled. We used a desktop equipped with an AMD Ryzen™ 9 5900X processor to execute these algorithms.

\hl{The parameter setting of the three methods is mainly determined by experience. We will tune the parameters for each method until it reaches the best results. The RANSAC L-shape method involves parameters about a pre-set threshold and RANSAC. The pre-set threshold to determine the shape of the point cloud is set to 30°. For the RANSAC, the minimum iteration number is 35, the maximum iteration number is 100 and the inlier threshold is 0.12. The iteration resolution of the CHH method and our method are both set to 0.5° to ensure fairness.
}

In order to exclude the influence of ground segmentation and clustering, we used the labeled information to extract the point cloud for every single object directly. Since we mainly focus on vehicle pose estimation, pedestrians and other non-vehicle objects in the KITTI dataset will not participate in the evaluation. Then the vehicle pose estimation algorithm will read the single object’s point cloud and calculate its orientation. It is worth noting that the orientation value in KITTI labels ranges from -$\pi$ to $\pi$. We convert it to the range of 0 to $\pi/2$ to be consistent with our algorithm. After comparing the estimated value with the ground truth information, we can evaluate the pose estimation performance.

For evaluating the accuracy of vehicle pose estimation, we utilized widely used evaluation metrics including orientation error and absolute orientation error \cite{ zhao2021shape,zhang2017efficient}. The absolute orientation error reflects the estimation bias without considering its direction. Furthermore, the running time was also measured to verify that our method is computationally efficient.

\subsection{KITTI dataset}
The KITTI \hl{3D object detection} dataset comprises 7481 training images, 7518 test images, and their corresponding point clouds. The point clouds were collected by a Velodyne HDL-64E rotating 3D LiDAR. A total of 26824 objects in the KITTI dataset will be evaluated.

\begin{table*}[h]
\caption{\label{tab:kitti_perm}Orientation error and running time comparison between different fitting methods on the KITTI dataset.}
\centering
\small
\begin{tabular}{@{}lllllll}
\br
&\centre{2}{Error (deg)}&\centre{2}{\hl{Absolute} Error (deg)}&\centre{2}{Running Time (ms)}\\
\ns
&\crule{2}&\crule{2}&\crule{2}\\

Methods &Mean \ $\downarrow$ &STD \ $\downarrow$ &Mean \ $\downarrow$ &STD \ $\downarrow$ &Mean \ $\downarrow$ &STD \ $\downarrow$\\
\mr
RANSAC L-shape & 1.5586 & 8.0333 & 3.8034 & 7.2455  & 0.3612 & 0.6333\\
CHH &  1.4815 & 7.1813 & 3.3648 & 6.5149  & \textbf{0.2413} & \textbf{0.2056}\\
Proposed & \textbf{0.5075} &  \textbf{4.2269} &\textbf{1.7299} & \textbf{3.8899} & 0.2872 & 0.2124\\
\br
\end{tabular}
\end{table*}

\begin{figure*}
\centering
\subfloat[RANSAC L-shape]{\label{edt-a} \includegraphics[scale=0.9]{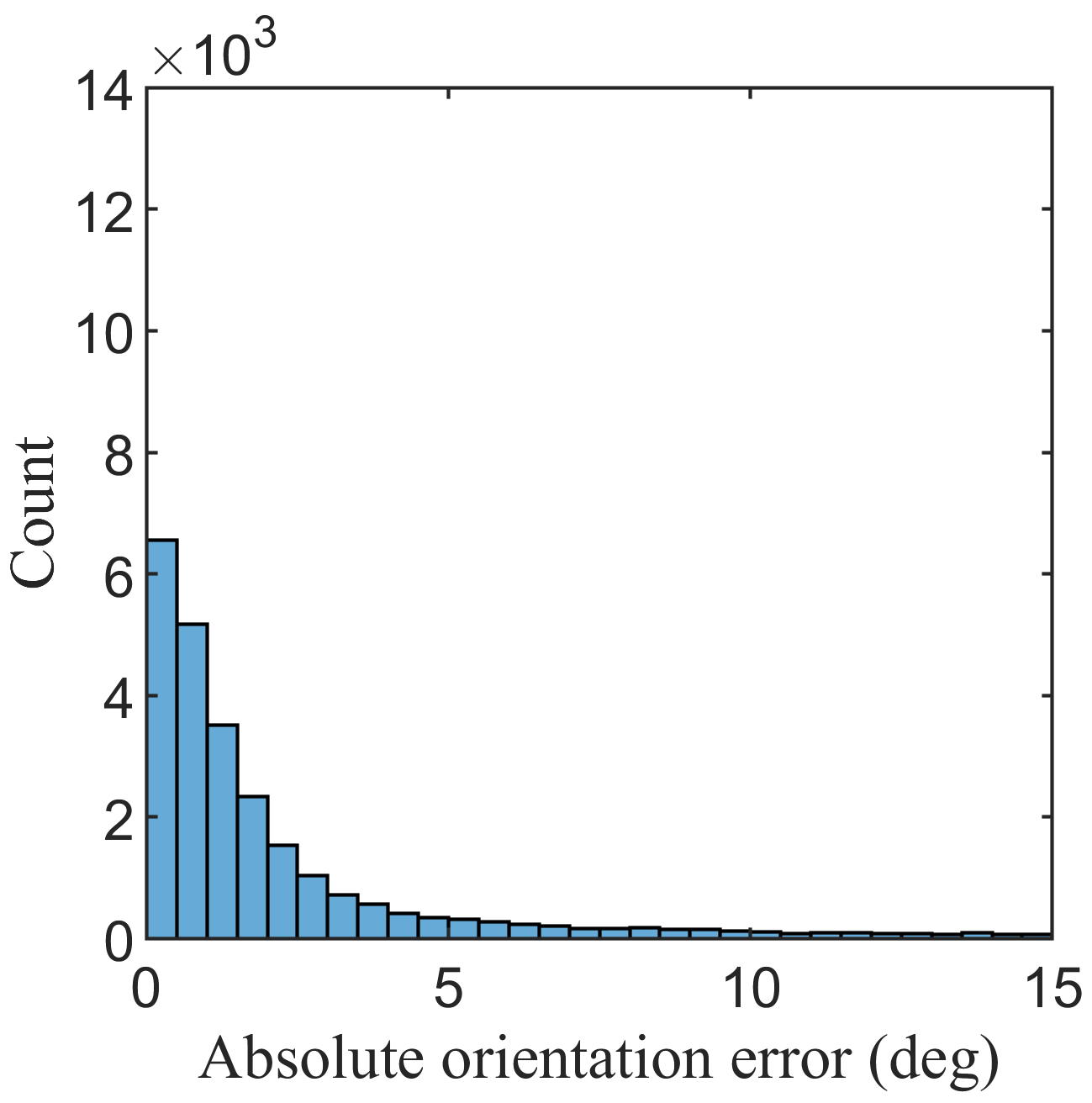}}
\hfill
\subfloat[CHH]{\label{edt-b} \includegraphics[scale=0.9]{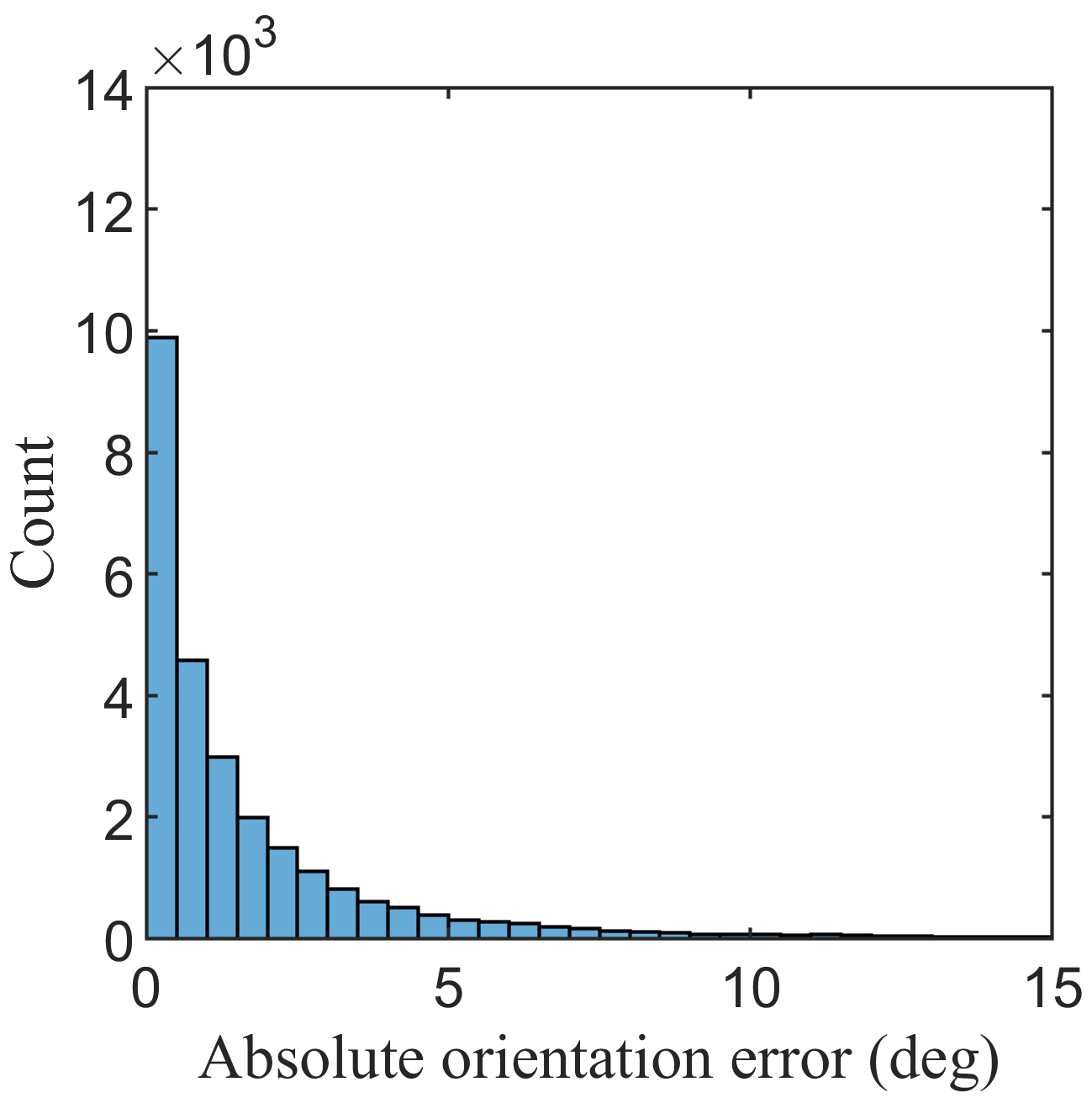}}%
\hfill
\subfloat[Proposed]{\label{edt-c} \includegraphics[scale=0.9]{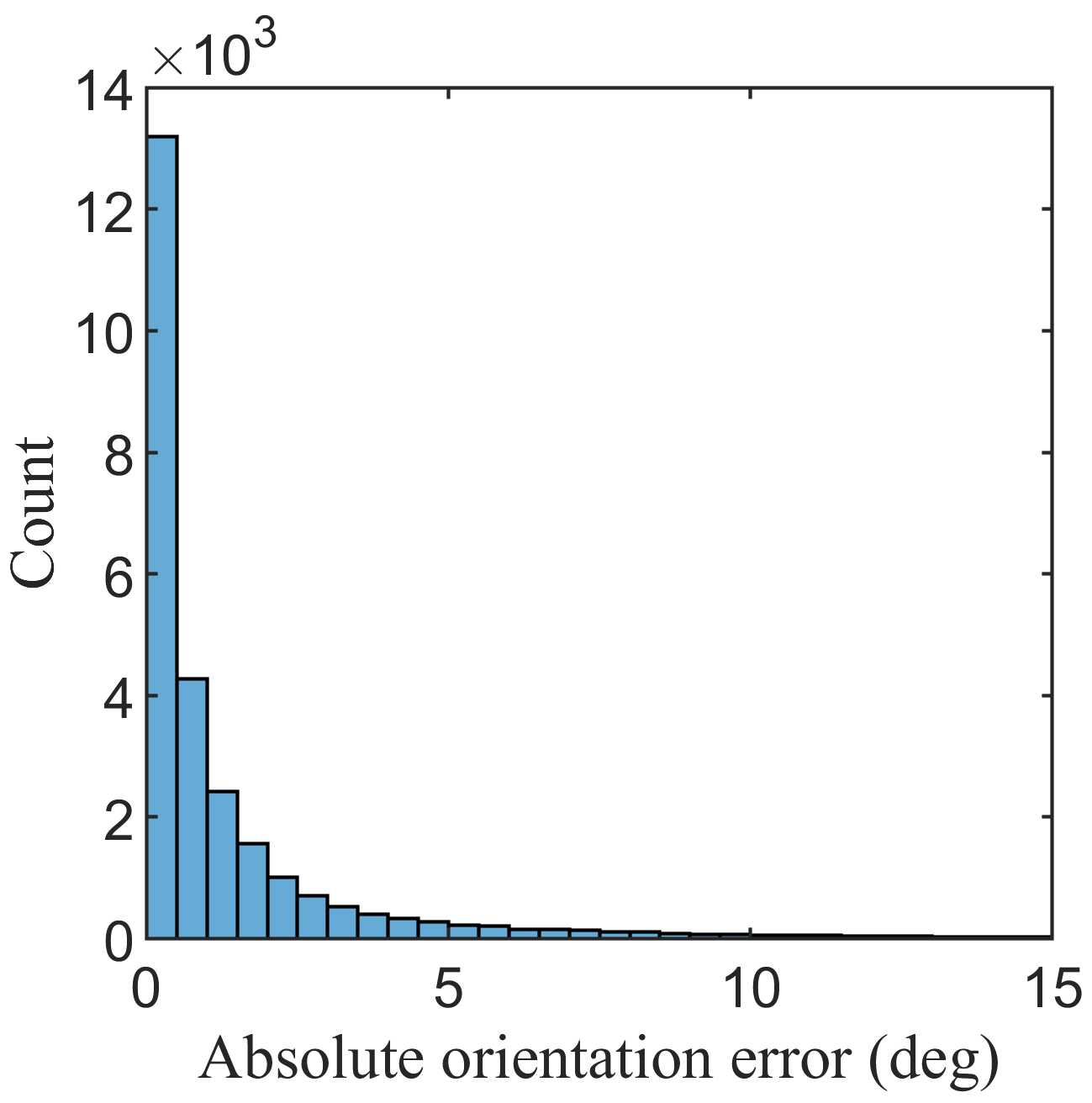}}%
\caption{Distribution of absolute orientation error on KITTI dataset}
\label{fig:error_distrib_kitti}
\end{figure*}

The mean and standard deviation of orientation error and absolute orientation error are listed in Table \ref{tab:kitti_perm}. The RANSAC L-shape method gets the worst result, especially in the STD item which shows that this method is not very stable. \hl{This is because this method relies on the edge features and a line-fitting algorithm.} The CHH method has a few improvements over the RANSAC L-shape method. Compared with the CHH method, the proposed algorithm improves the mean absolute orientation error by 48.6\% from 3.3648° to 1.7299°, which shows our method achieves significant improvement in vehicle pose estimation. Besides, our method also has a 40\% improvement in STD items which proves our method is more robust. Table \ref{tab:kitti_perm} also presents the computation time fitting one object. All three methods achieve good computation efficiency. The proposed method only spends an average of 0.2872 ms for one single object, which can achieve real-time performance even fitting 100 objects simultaneously. Figure \ref{fig:error_distrib_kitti} illustrates the distribution of absolute orientation error on the KITTI dataset. As shown in Figure \ref{fig:error_distrib_kitti}, most errors are distributed from 0° to 1° with the proposed method. On the contrary, the other two methods have much fewer results from 0° to 1° and have a more scattered distribution. These results suggest that the proposed method is more consistent and stable than other methods.

\begin{figure}[tb]
    \centering
    \subfloat[\label{1a}][Case 1]{
        \includegraphics[scale=0.12]{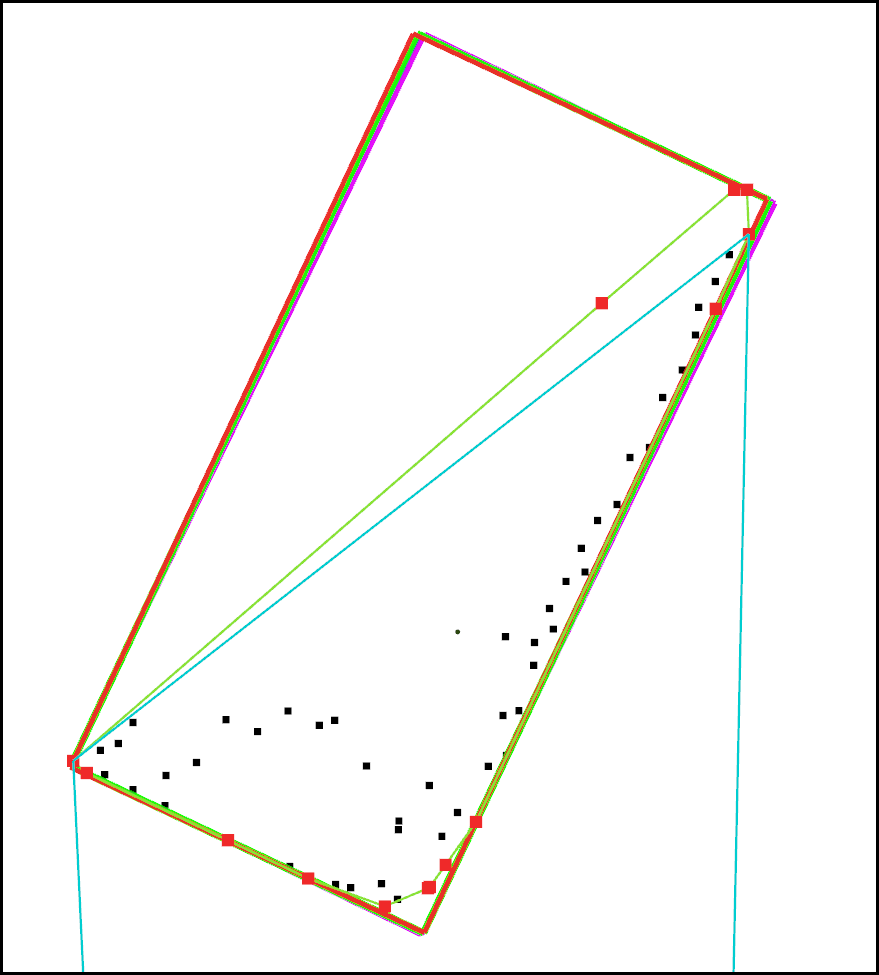}}
    \subfloat[\label{1b}][Case 2]{
        \includegraphics[scale=0.12]{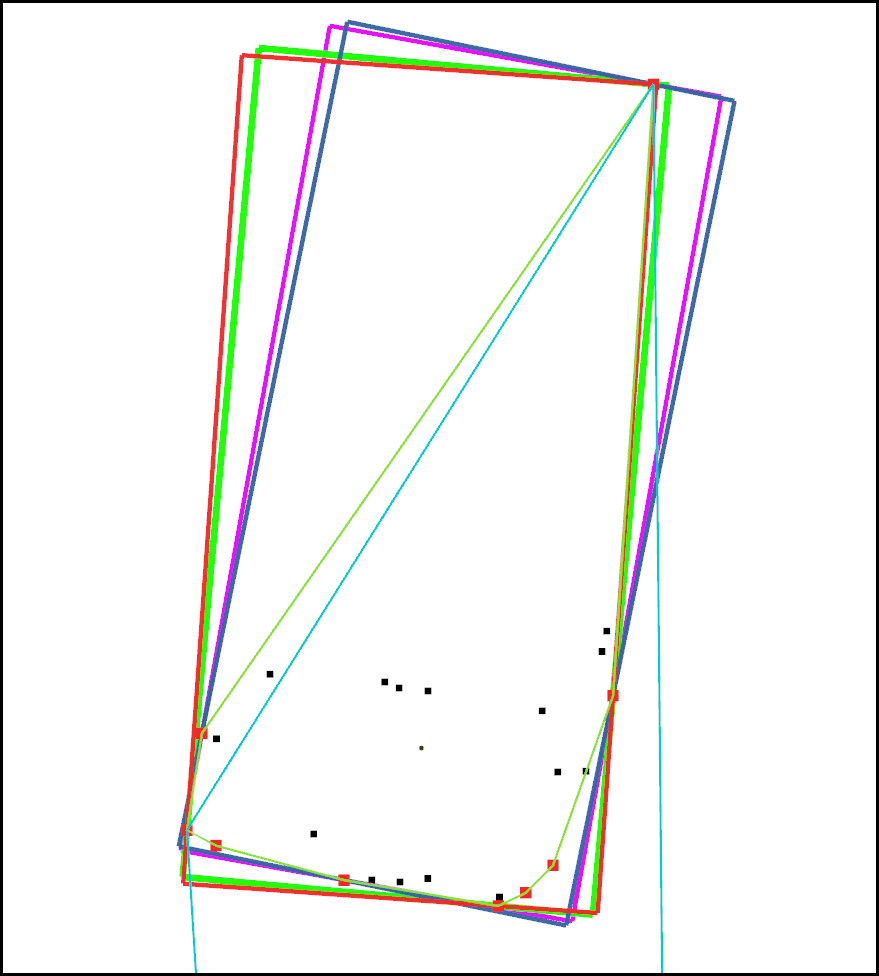}}\\
    \subfloat[\label{1c}][Case 3]{
        \includegraphics[scale=0.12]{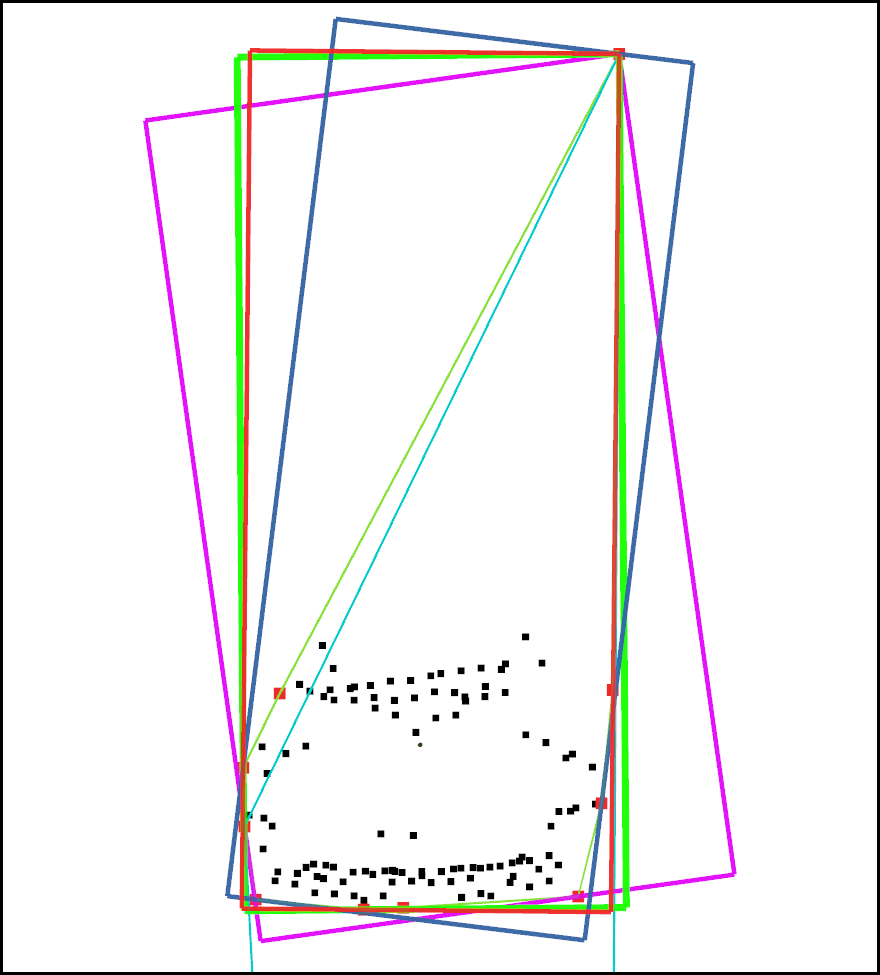}}
    \subfloat[\label{1d}][Case 4]{
        \includegraphics[scale=0.12]{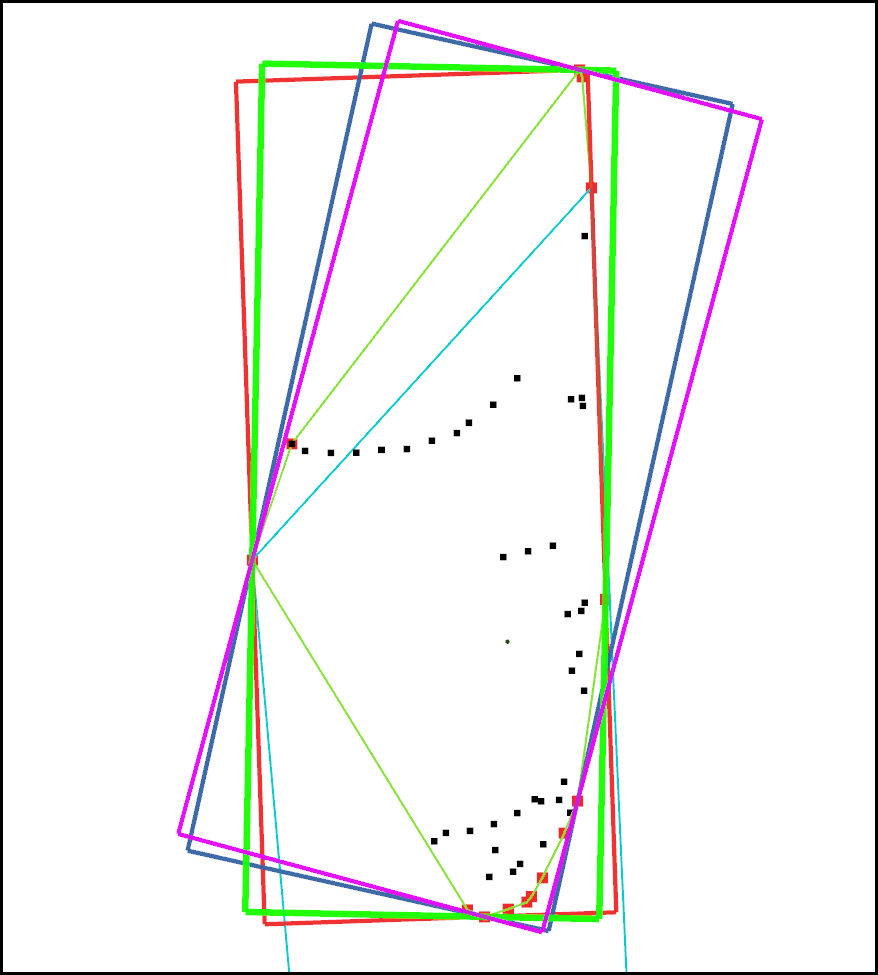}}
	\caption{Sample results of vehicle pose estimation on KITTI dataset. Green boxes from the ground truth, purple boxes from the RANSAC L-shape fitting method, blue boxes from the CHH method, and red boxes from the proposed algorithm.}
    \label{fig:sample_kitti}
\end{figure}

The examples of fitting results on the KITTI dataset are shown in Figure \ref{fig:sample_kitti}. In case 1, the cluster is L-shape which is very common in the KITTI dataset, and all three methods get a good fitting result. The cluster in case 2 is still an L-shape but very sparse. Only the proposed method achieves good fitting. \hl{The RANSAC L-shape method is unable to fit a line accurately using such a sparse point cloud. The CHH method is because the closeness criterion score reaches a minimum in the wrong pose.} In case 3, the cluster is occluded by the front of the vehicle. Our method achieves the best-fitting result, and the other two methods have large orientation estimation errors. \hl{The RANSAC method results in a false fitting due to two curved point clouds at the front. The CHH method is still limited to the closeness criterion.} The cluster in case 4 has occlusion. The proposed method can still get the right fitting result. \hl{The RANSAC L-shape method gets the wrong fitting due to the contour of the point cloud is not standard L-shaped or I-shared. The CHH method remains constrained by the closeness criterion.}

\subsection{Own dataset}
\begin{figure}[t]
\centerline{\includegraphics[scale=0.23]{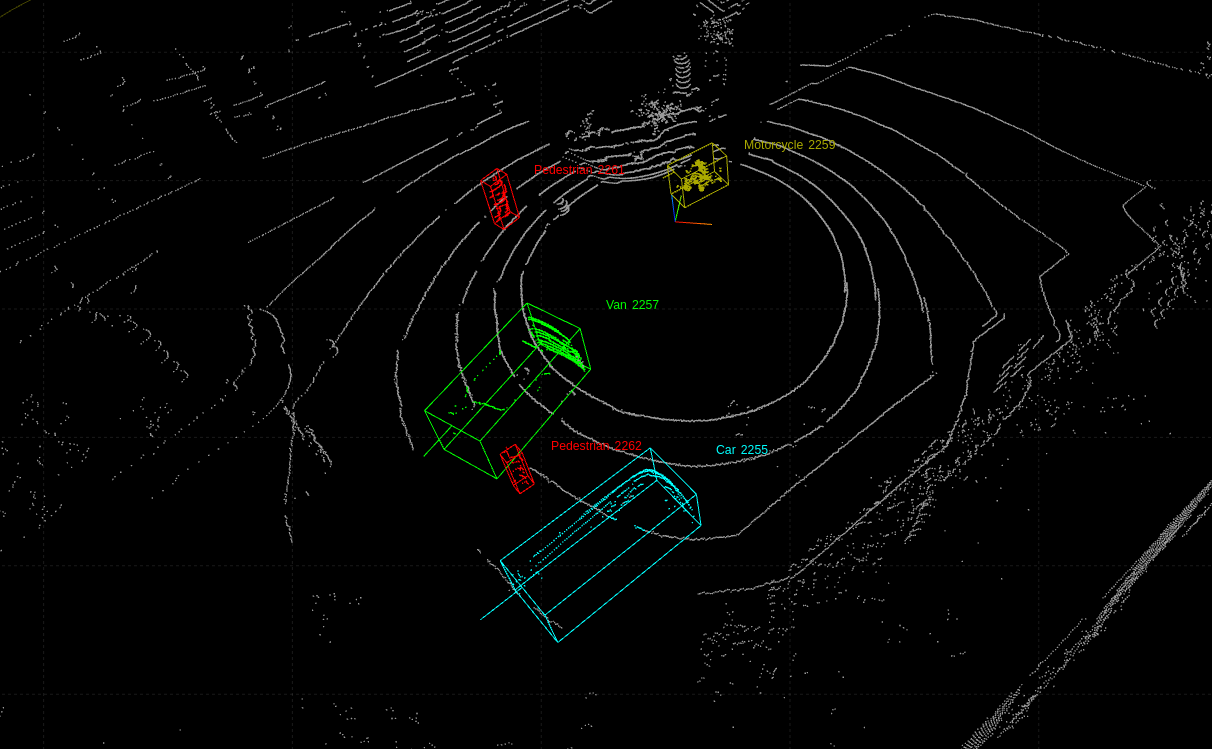}}
\caption{A demonstration of our own dataset and its annotation.
\label{fig:self_clollected}}
\end{figure}

Our own dataset is collected with a Velodyne Puck (VLP-16) 3D LiDAR in an industrial park, as shown in Figure \ref{fig:self_clollected}. We labeled 600 frames of LiDAR point cloud using the 3D bounding box annotation tool: SUSTechPOINTS \cite{li2020sustech}. Then we convert these labels to the KITTI format. The rest procedures are the same as the KITTI dataset.

\begin{figure*}
\centering
\subfloat[RANSAC L-shape]{\label{3figs-a} \includegraphics[scale=0.9]{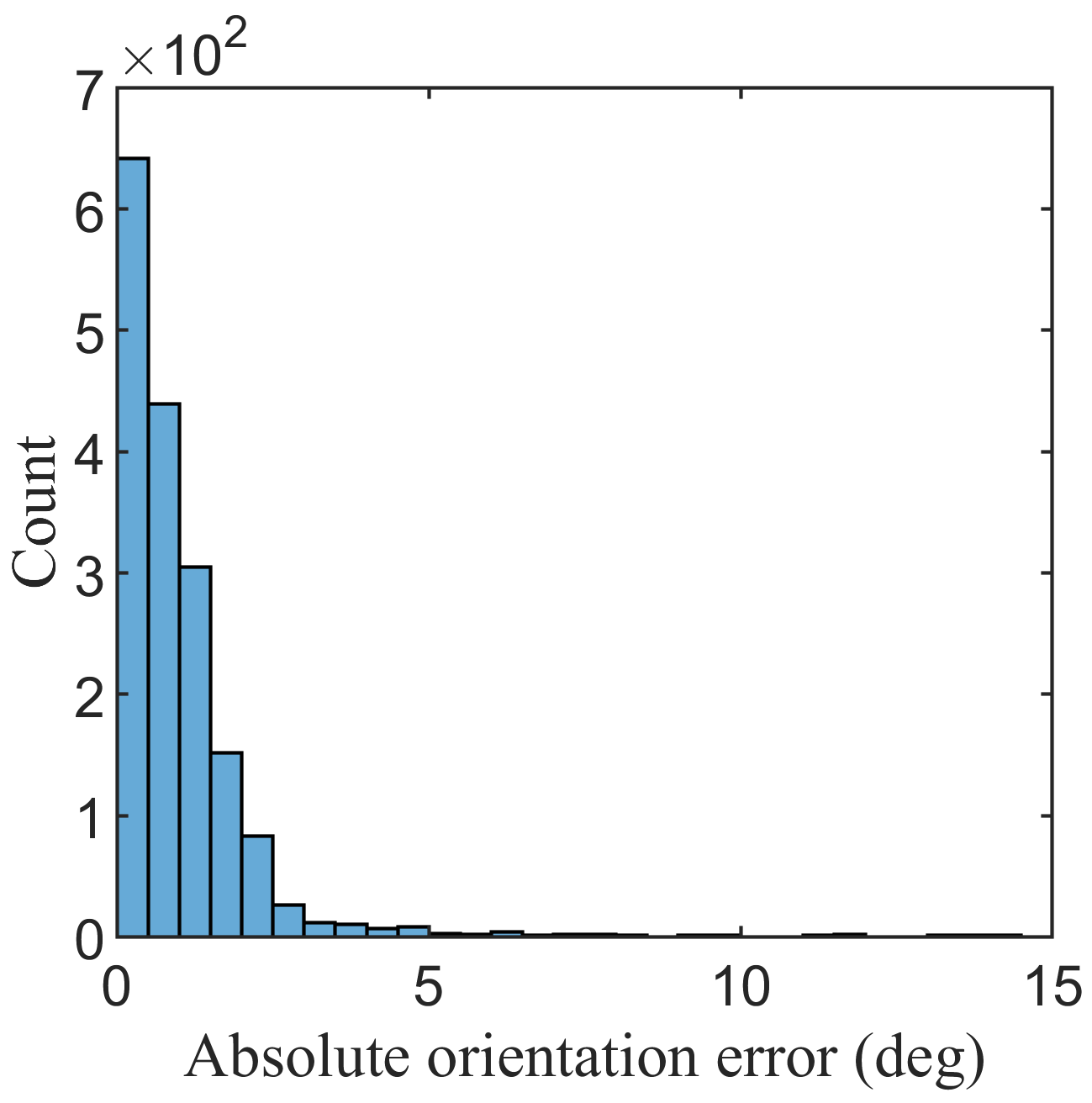}}
\hfill
\subfloat[CHH]{\label{3figs-b} \includegraphics[scale=0.9]{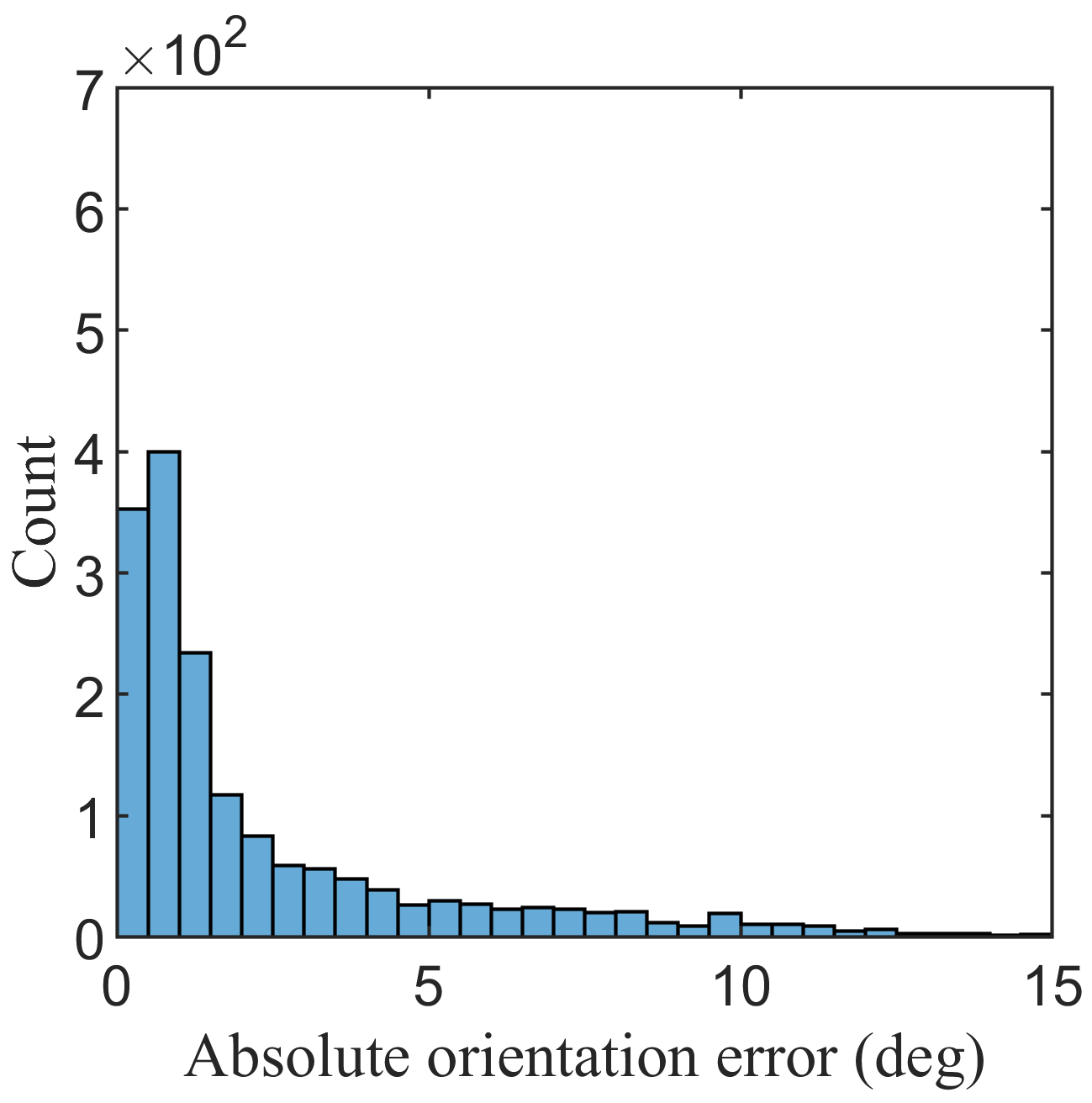}}%
\hfill
\subfloat[Proposed]{\label{3figs-c} \includegraphics[scale=0.9]{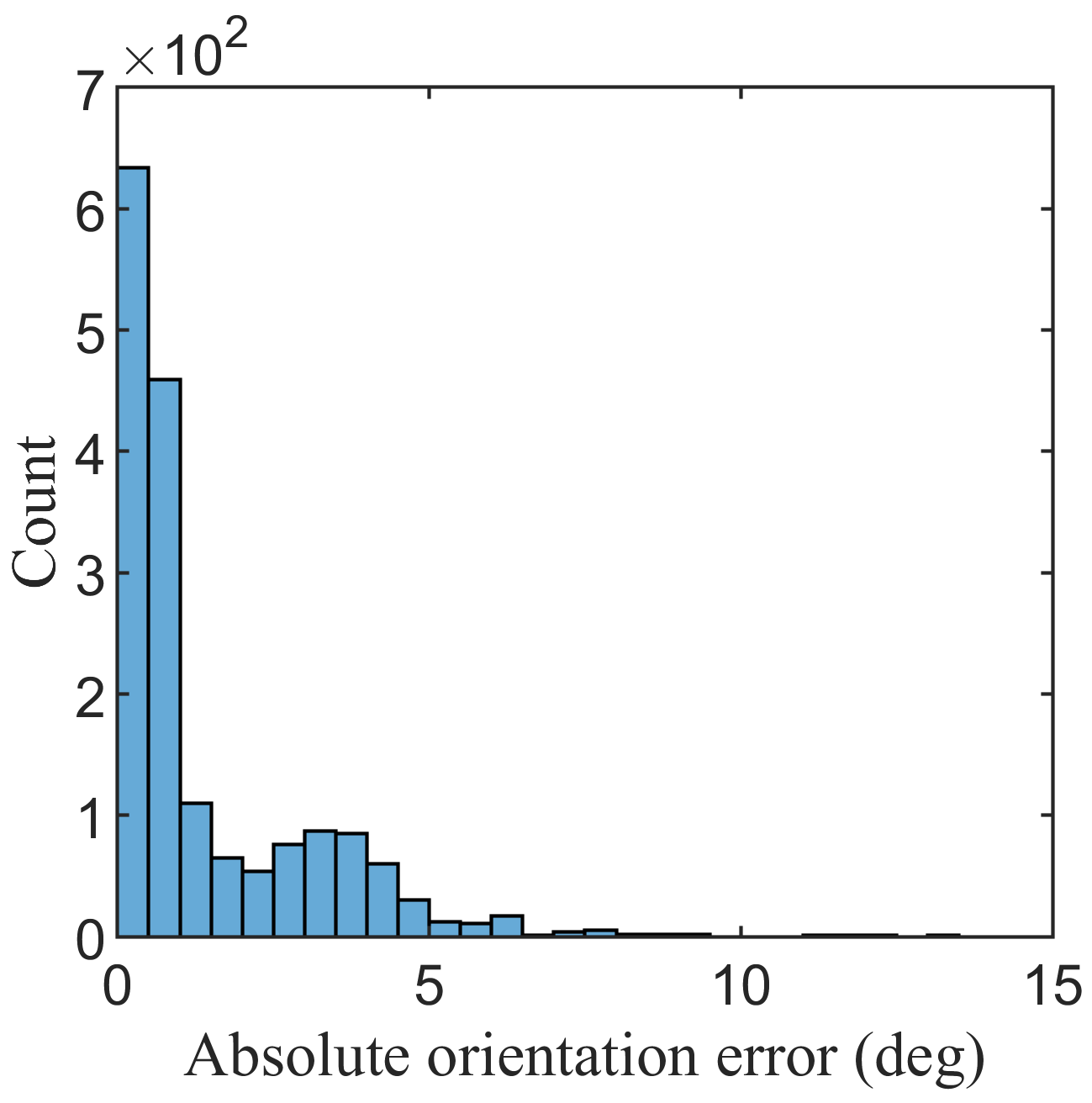}}%
\caption{Distribution of absolute orientation error on our own dataset.}
\label{fig:error_distrib_own}
\end{figure*}

Table \ref{tab:own_perm} presents the quantization results of pose estimation on our own dataset. Compared to the KITTI dataset, the mean and STD of orientation error dropped a lot, and the running time increased. This is because the objects we labeled are located at a relatively close distance. Therefore, the point cloud of the labeled objects is denser and less occluded. For the RANSAC L-shape fitting method, its pose estimation accuracy improves a lot, only slightly worse than our algorithm. However, its running time has almost tripled compared with the KITTI dataset, which shows that this method is sensitive to the number of point clouds. The proposed algorithm still gets the best result on our own dataset except for the running time. Figure \ref{fig:error_distrib_own} presents the distribution of absolute orientation error on our own dataset. As shown in Figure \ref{fig:error_distrib_own}, the proposed method and RANSAC L-shape fitting method have a more concentrated distribution. However, there is a peak near 4° in the histogram of our method. \hl{This is because the collected data contains many vehicles that have rearview mirror point clouds. Our method may not achieve good estimation results in such cases.}

\begin{table*}[h]
\caption{\label{tab:own_perm}Orientation error and running time comparison between different fitting methods on our own dataset.}
\centering
\small
\begin{tabular}{@{}lllllll}
\br
&\centre{2}{Error (deg)}&\centre{2}{\hl{Absolute} Error (deg)}&\centre{2}{Running Time (ms)}\\
\ns
&\crule{2}&\crule{2}&\crule{2}\\

Methods &Mean \ $\downarrow$ &STD \ $\downarrow$ &Mean \ $\downarrow$ &STD \ $\downarrow$ &Mean \ $\downarrow$ &STD \ $\downarrow$\\
\mr
RANSAC L-shape & 0.2441 & 3.8377 & 1.3325 & 3.6072 & 1.0656 & 1.0650\\
CHH & 1.1150 & 6.2983 & 2.9992 & 5.6494 & \textbf{0.3372} & \textbf{0.1295}\\
Proposed & \textbf{-0.0115} & \textbf{2.5989} & \textbf{1.3101} & \textbf{2.2445}  & 0.3763 & 0.1406\\
\br
\end{tabular}
\end{table*}

\begin{figure}[tb]
    \centering
    \subfloat[\label{2a}][Case 1]{
        \includegraphics[scale=0.12]{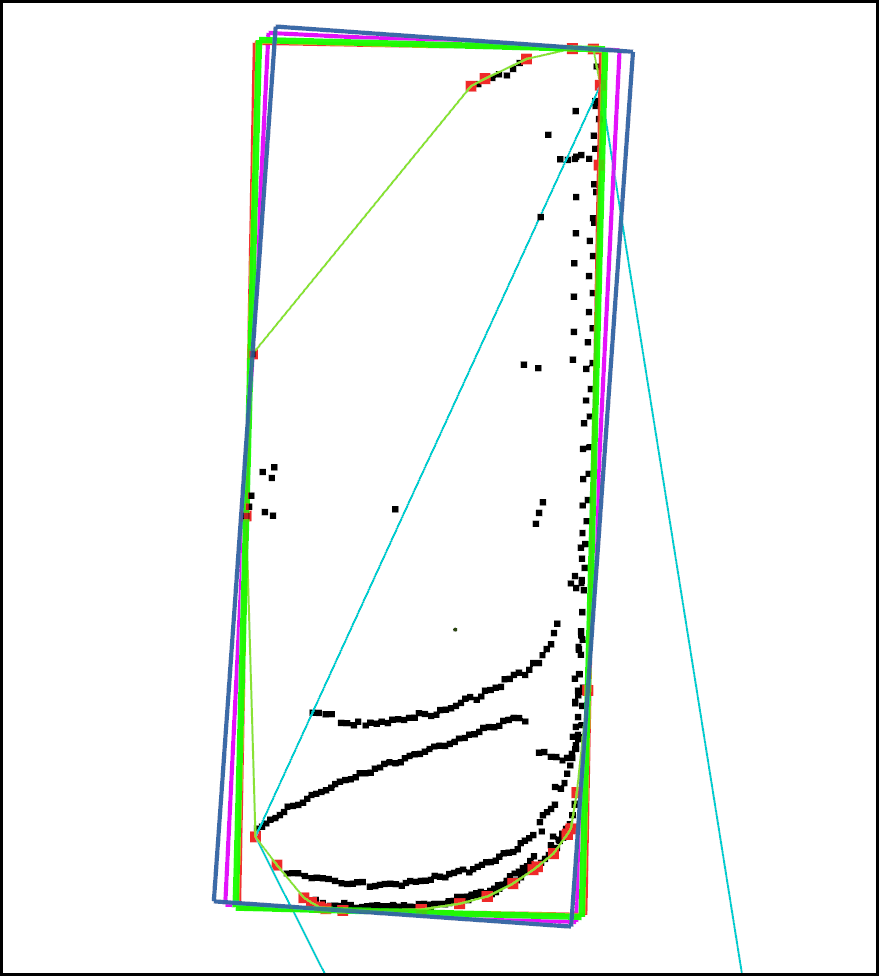}}
    \subfloat[\label{2b}][Case 2]{    
        \includegraphics[scale=0.12]{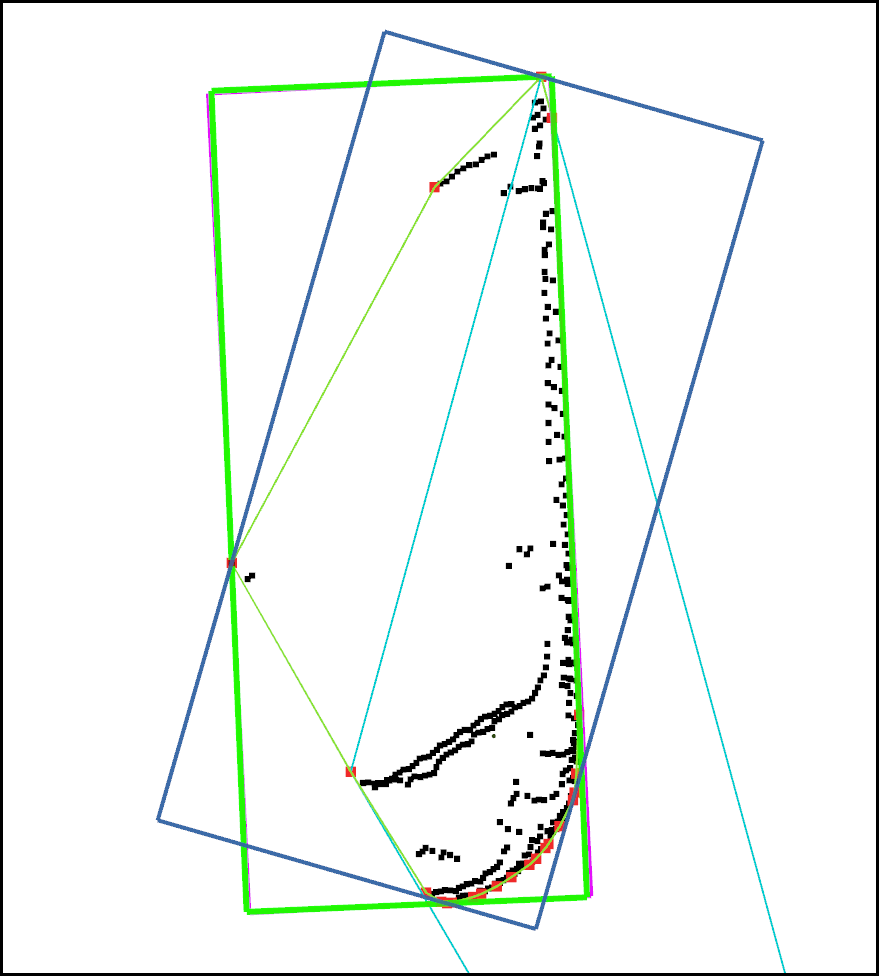}}\\
    \subfloat[\label{2c}][Case 3]{
        \includegraphics[scale=0.12]{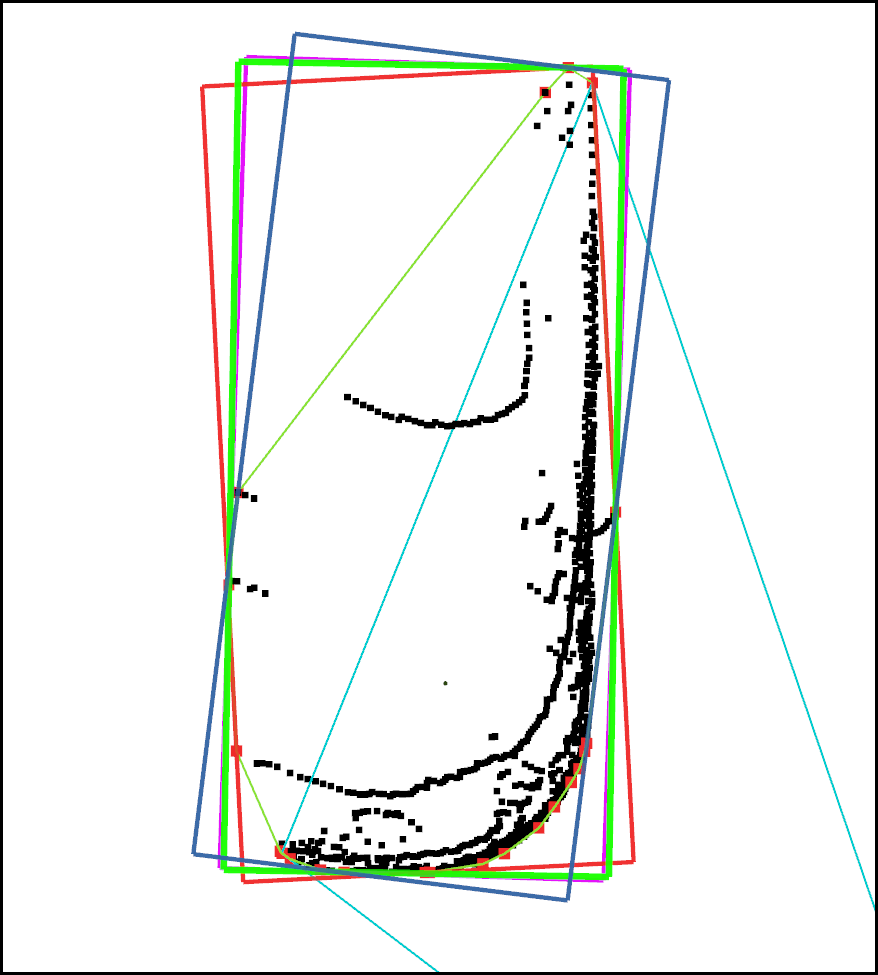}}
    \subfloat[\label{2d}][Case 4]{
        \includegraphics[scale=0.12]{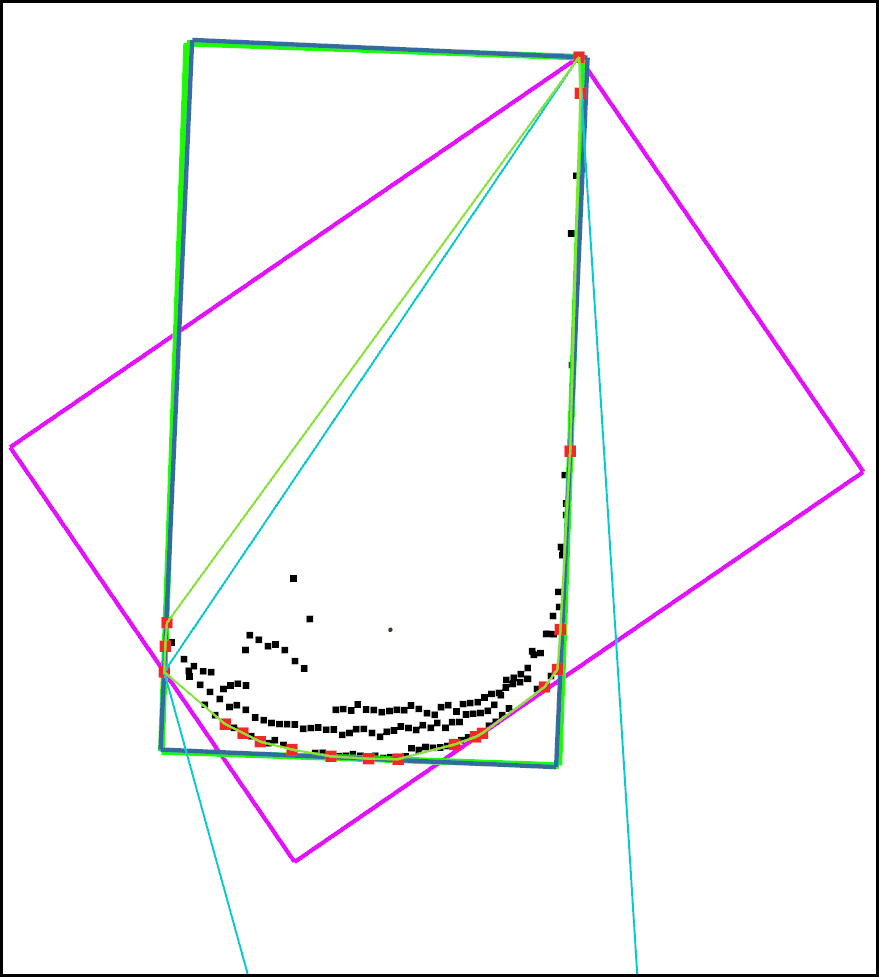}}
    \caption{Sample results of vehicle pose estimation on our own dataset. Green boxes from the ground truth, purple boxes from the RANSAC L-shape fitting method, blue boxes from the CHH method, and red boxes from the proposed algorithm.}
    \label{fig:sample_own}
\end{figure}

The examples of fitting results on our own dataset are shown in Figure \ref{fig:sample_own}. In case 1, the cluster is L-shaped which is also common in our own dataset. All three methods achieve correct fitting, but our method gets the best-fitting result. The cluster in case 2 exists occlusion. \hl{The CHH method gets the wrong orientation estimation due to the CHH method having the highest score on the closeness criterion after convex hull extraction.} The other two methods get a good fitting. In case 3, there exists a rear-view mirror on the side of the vehicle, which can be seen often in our dataset. Only the RANSAC L-shape fitting method achieves good fitting. \hl{Both our method and the CHH method exhibit sensitivity to protruding objects, stemming from the inclusion of this feature during convex hull extraction.} This phenomenon elucidates the peak near 4° observed in the histogram of our method. In case 4, the vehicle is occluded by the front of the vehicle, and there are only a few points on the side of the vehicle. The RANSAC L-shape fitting method gets the wrong fitting due to the cluster being curving. The other two methods achieve good fitting.

\section{Conclusions}
A novel convex hull-based vehicle pose estimation method is introduced in this paper. 
\hl{To improve computational efficiency, we employ convex hull extraction to obtain the contour information. Then the visible edge selection module is used to determine the projection edges. Finally, a novel criterion based on the minimum occlusion area is developed to get the vehicle pose estimation for various point cloud distributions.} Experimental results on the KITTI dataset and a manually labeled dataset show that the proposed method can estimate the vehicle's pose accurately and stably while maintaining real-time speed. In our future work, we aim to incorporate our method with tracking algorithms to smooth the orientation estimation result and reduce unnecessary searches. \hl{Considering the proposed algorithm is sensitive to protruding objects or outlier noisy points, we want to introduce RANSAC in our method to solve this limitation.}


\section*{Conflicts of interest}
The authors have no relevant financial or nonfinancial interests to disclose

\section*{Data availability statement}
The data generated and/or analyzed during the current study are not publicly available for legal/ethical reasons but are available from the corresponding author on reasonable
request.




\section*{References}
\bibliographystyle{iopart-num-long}
\bibliography{main.bib}

\newpage{}
\appendix

\end{document}